\documentclass[11pt,a4paper]{article}
\usepackage[hyperref]{acl2020}
\usepackage{times}
\usepackage{latexsym}

\usepackage{microtype}

\usepackage{graphicx}
\usepackage{amssymb}
\usepackage{amsmath}
\usepackage{mathtools}
\usepackage{booktabs}

\mathchardef\mhyphen="2D 

\usepackage{standalone}
\usepackage{tikz}
\usepackage{pgfplots}
\pgfplotsset{compat=1.14}
\usepackage{booktabs}

\pgfplotscreateplotcyclelist{tb}{
semithick,orange,mark=+\\%
semithick,blue,mark=+\\%
semithick,red,mark=+\\%
semithick,cyan,mark=+\\%
}

\aclfinalcopy 

\title{Image Captioning with Visual Object Representations Grounded \\ in the Textual Modality}

\author{
    Du\v{s}an Vari\v{s}$^1$\thanks{* Contribution during the student internship at RIKEN AIP.}\footnotemark[1], Katsuhito Sudoh$^{2,3}$, Satoshi Nakamura$^{2,3}$ \\
    $^1$Charles University, Faculty of Mathematics and Physics, Czechia \\
    $^2$Nara Institute of Science and Technology, Japan\\
    $^3$RIKEN AIP, Japan \\
    \texttt{varis@ufal.mff.cuni.com} \\
    \texttt{\{sudoh,s-nakamura\}@is.naist.jp}}

\date{}

\begin{document}
\maketitle
\begin{abstract}
    We present our work in progress exploring the possibilities of a shared embedding space between textual and visual modality.
    Leveraging the textual nature of object detection labels and the hypothetical expressiveness of extracted visual object representations, we propose an approach opposite to the current trend, grounding of the representations in the word embedding space of the captioning system instead of grounding words or sentences in their associated images.
    
    
    Based on the previous work, we apply additional grounding losses to the image captioning training objective aiming to force visual object representations to create more heterogeneous clusters based on their class label and copy a semantic structure of the word embedding space.
    In addition, we provide an analysis of the learned object vector space projection and its impact on the IC system performance.
    
    
    With only slight change in performance, grounded models reach the stopping criterion during training faster than the unconstrained model, needing about two to three times less training updates.
    Additionally, an improvement in structural correlation between the word embeddings and both original and projected object vectors suggests that the grounding is actually mutual.
    
    
    

\end{abstract}

\section{Introduction}
    The task of image captioning (IC) or image description generation focuses on generating a sentence describing a given input image.
    Combining the tasks from computer vision and natural language generation, IC has become an important task for the currently dominant deep learning research.
    The emergence of image captioning datasets such as MSCOCO \cite{Lin2014mscoco} or Flick30k \cite{plummer2017flickr30k} has also greatly helped the research progress in this area.
    
    Throughout the years, several captioning approaches have been suggested.
    In the early days of neural IC, the systems were trained mainly in the end-to-end fashion using an output of a convolutional neural network (CNN) for conditioning a sequential language model to generate an output caption.
    The CNN output consisting of some mid-level image features was either used to initialize the language model decoding \cite{vinyals2015show} or fed directly into each decoding step via attention mechanism \cite{xu2015showattend}.
    Despite its good performance, these models are hard to interpret and it has been shown that they mainly exploit distributional similarity in the multimodal feature space \cite{madhyastha2018exploits}.
    
    Recently, several methods that suggest using information about the objects present in the image were revisited.
    Object information is usually extracted via an object detection network \cite{girshick2014rcnn, shaoqing2015fasterrcnn,redmon2016yolo} and consists of a set of bounding boxes locating the objects and labels that describe them.
    Treating such labels as a simple bag-of-objects (BOO) representations \citet{wang2018counts} or using a context-dependent encoding of the detected objects \citet{yin2017obj2text} and using them to initialize the language model can outperform IC systems using CNN features as their input.
    Also, using explicitly detected objects makes these models easier to interpret.
    
\begin{figure}[!tbp]
    \centering
    \begin{minipage}[b]{0.9\linewidth}
        \centering
        \includegraphics[width=\textwidth]{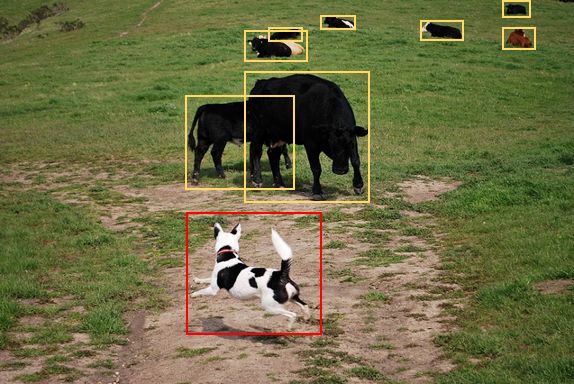}
    \end{minipage}

    \begin{minipage}[b]{0.9\linewidth}
       \centering
        \includegraphics[width=\textwidth]{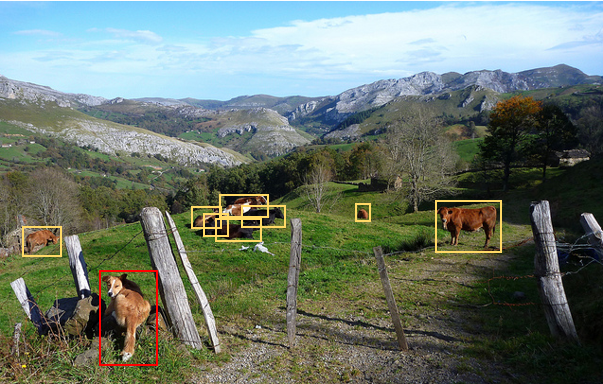}
    \end{minipage}
    \caption{Example of two images where an identical sets of objects were detected even though the actions related to the objects differ (e.g. upper: dog jumping, lower: dog lying down).}
    \label{fig:motivation}
\end{figure}
    
    Even though the detection-based IC systems show promising results, by design, they are bound to fail at distinguishing images with identical BOO representations as shown in Figure~\ref{fig:motivation}.
    The image contains the same number of detected animals (dogs and cows) with different location distribution. 
    Even though we can alleviate the BOO identity by incorporating the location \cite{yin2017obj2text} and size \cite{wang2018counts} of the objects into the model, it should be impossible for the model to capture details about the actions tied to the objects (dog jumping vs dog lying down).
    Using a more robust (detailed) object detector might help to solve this problem, but we argue that it is often too costly and task-specific.
    
    Instead of using embedded object detection labels, \citet{anderson2017up-down} proposed a method using intermediate object representations (called bottom-up attention) extracted from the CNN features via their object bounding boxes. According to their results, these object features are more effective than the ``raw'' CNN features.
    We also hypothesise that these features are more descriptive than the combination of object labels and bounding boxes.
    The potential problem of this approach is that the parts of the network (CNN, object detection network) are trained independently which can lead to different representational spaces for each modality.
    This can be partially fixed by end-to-end fine-tuning of the whole captioning pipeline, but again this can be quite costly.
    
    In this work we propose a novel method of grounding of the pretrained intermediate visual object representations (VOR) in the textual embedding space. We also focus on the analysis of the learned VOR projections and their relationship with the IC captioning performance. The paper is structured as follows: In section~\ref{sec:relwork}, we provide a brief overview of the previous work on multimodal grounding and shared embedding space representational learning. In section~\ref{sec:proposed}, we describe our method. In section~\ref{sec:experiments}, we describe the settings and results of our experiments and the learned projection space analysis. Section~\ref{sec:conclusion} concludes the paper.
    
\section{Related Work}
\label{sec:relwork}
    
    Currently, there is an ongoing research studying shared multimodal representation learning.
    Inspired by human perception, the aim is to find a way to comprehensively represent information from various sensory inputs.
    In the case of deep learning, these sensory inputs are usually represented by different parts of the neural network model.
    This is often difficult to accomplish either due to a fact that we usually train these parts of the network  separately on different tasks or by a lack of constraints that would force the network to learn better semantically correlated representations between the modalities, the latter being referred to as a heterogeneity gap \cite{guo2019replearningsurvey}.
    
    In the case of combining a visual and textual modality, current research focuses mainly on the natural language grounding in its visual counterpart.
    This is motivated by a difference in occurrence between the textual representation of objects and their relations and their actual real-world frequency \cite{gordon2013textbias}.
    Majority of the current work therefore focus on reducing this bias between text and reality.
    
    There are generally two approaches towards the textual grounding: a word-level and a sentence-level approach.
    The word-level methods either learn grounded representations separately for each modality first and combine them later \cite{silberer2014learning,collell2017imagined} or learn them jointly from multiple sources \cite{hill2016learning}.
    This can be done by taking a standard word embedding objective, e.g. continuous bag-of-words \cite{mikolov2013efficient} or skip-gram \cite{mikolov2013word2vec} and introducing an additional training objective to force the word vectors to imitate the structure of the related visual features \cite{lazaridou2015combining} or visual context \cite{zablocki2018learning}.
    
    The sentence-level methods usually try to learn grounded sentence representations by adding an additional training objective for predicting visual features associated with the sentence \cite{chrupala2015learning, kiela2018learning}.
    They accomplish this via cross-modal projection, however, according to \citet{collell2018crossmodal}, the learned cross-modal mapping does not properly preserve the structure of the original representation space.
    To address this issue, \citet{bordes2019groundedspace} suggest using an intermediate grounded space as it weakens the imposed constraint on the textual representations.

    The currently most common approach towards IC is to condition a language model on the intermediate visual features extracted by an image processing network.
    These features can be either general feature map or bounded to objects detected inside the picture.
    Ideally, we would like the language model to extract important details about the input from these features, however, \citet{madhyastha2018exploits} show that
    these models mostly exploit a distributional similarity in the multimodal feature space, finding the closest training examples in the multimodal space and combining their captions to describe the test image.
    
    In spite of that, we think that intermediate VORs extracted by an object detection network can contain more useful information than we are currently able to exploit.
    Since word embeddings are a crucial part of the IC decoder, we assume that by grounding object features in the word embedding space using the embeddings of their labels can help the IC decoder to better extract structural information about the input image.
    This idea seems even more plausible with the attention-based IC decoders \cite{xu2015showattend, anderson2017up-down}, where decoder learns to shift its focus to different parts of an image during the caption generation process.

    
    
    
\section{Proposed Method}
\label{sec:proposed}
    
    In this section, we describe our proposed method starting from its model architecture, the motivation behind grounding of VORs in a word embedding space and the investigated grounding methods.
    
    \subsection{Model}
    
    Our model implementation is based on the one proposed by \citet{anderson2017up-down}. For a given image, the model takes an input set of $k$ feature vectors $\textbf{v} = \{v_1 .. v_k\}$, $v_i \in \mathbb{R}^{d_{in}}$, where $d_{in}$ is the vector dimension.
    These vectors can be extracted either directly from the object detection network, such as Faster RCNN \cite{shaoqing2015fasterrcnn} or by cropping out the relevant object regions from the CNN feature map using methods such as region-of-interest (RoI) pooling \cite{girshick2014rcnn}.
    
    Since we use only fixed image/object feature vectors in this work, we want to learn a following transformation $f(\cdot)$ of the input:
    \begin{equation}
        z_{i} = f(v_{i})
    \end{equation}
    where $z_{i} \in \mathbb{R}^{d}$, $d$ being a dimension of the projected vector space, is a feature representation specific for our image captioning task.
    The transformation $f(\cdot)$ can be either linear transformation or a MLP.
    
\begin{figure}[!tbp]
    \centering
    \includegraphics[width=0.45\textwidth]{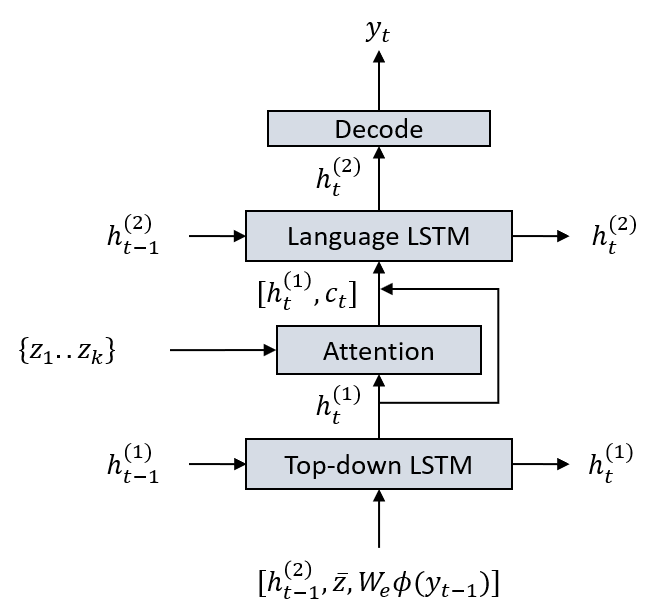}
    \caption{Overview of the the IC decoder architecture used during our experiments, similar to \citet{anderson2017up-down}. Instead of using pretrained feature vectors ${\textbf{v} = \{v_1 .. v_k\}}$ directly we use their trainable projection $\textbf{z} = \{z_1 .. z_k\}$.}
    \label{fig:architecture}
\end{figure}
    
    The transformed features $\textbf{z} = \{z_1 .. z_k\}$ are then passed to the LSTM decoder \cite{hochreiter1997lstm}.
    The decoder consists of two stacked LSTM layers; a top-down attention layer and a language model layer.
    Figure~\ref{fig:architecture} shows a schematic overview of the model.

    At time step $t \in \{1 .. T\}$, we encode a previously generated output symbol $y_{i-1}$ using an embedding matrix $W_e \in \mathbb{R}^{d\times |V|}$, where $V$ is the captioning symbol vocabulary, producing an input embedding vector $x_t$:
    \begin{equation}
        x_t = W_e\phi(y_{t-1})
    \end{equation}
    where $\phi$ converts the symbol $y_{i-1}$ to its corresponding one-hot encoding.
    We then pass a concatenation of the input vector $x_t$, mean-pooled transformed features $\overline{z} = \frac{1}{k}\sum_{i}z_i$, and a previous hidden state from the second layer $h_{t-1}^2$ to the top-down attention LSTM layer:
    \begin{equation}
        h_{t}^{1} = LSTM^1([x_t, \overline{z}, h_{t-1}^2], h_{t-1}^1)
    \end{equation}
    
    The top-down LSTM output $h_{t}^1$ is then passed to an attention layer together with the transformed feature vectors producing a context vector $c_t$:
    \begin{equation}
        c_t = Att(h_{t}^1, \textbf{z})
    \end{equation}
    where $Att(\cdot)$ is an attention mechanism \cite{bahdanau2015attention}.
    Following the previous work, we use the \textit{concat} attention described in \citet{luong2015attention}.
    
    A concatenated output of the attention layer $c_t$ and the top-down LSTM layer $h_t^1$ is the passed to the language LSTM layer producing a hidden state $h_t^2$:
    \begin{equation}
        h_t^2 = LSTM^2([c_t, h_t^1], h_{t-1}^2)
    \end{equation}
    Using the second hidden state, we produce a conditional probability distribution over the symbol vocabulary using a following transformation:
    \begin{equation}
        p(y_t | y_{1:t-1}) = Softmax(W_oh_t^2 + b_o)
    \end{equation}
    where $W_o \in \mathbb{R}^{|V|\times d}$ is output projection matrix and $b_o \in \mathbb{R}^{|V|}$ is bias.
    Getting the output probability distribution for each decoding time step, we can compute the distribution over the entire caption space as their product over the time steps:
    \begin{equation}
        p(y_{1:T}) = \prod_{t=1}^Tp(y|y_{1:t-1})
    \end{equation}
    
    We initialize $y_{0}$ with special beginning of sentence symbol and $h_0^1$, $h_0^2$ with zero values.
    
    \subsection{Grounding methods}
    
    The main idea behind the VOR grounding is to transform the object features in such a way that the IC decoder can extract necessary information in an easier way. This approach improves the performance on downstream tasks when the learned projection is between two spaces with a same modality \cite{artetxe2016learning, artetxe2018unsupervised}, however to our knowledge, it is still unclear how well this can be applied when different modalities are involved.
    
    We decided to perform an initial experiment with a simple IC system.
    We used MSCOCO \cite{Lin2014mscoco} with Karpathy splits \cite{karpathy2015deep}\footnote{\url{http://cs.stanford.edu/people/karpathy/deepimagesent}}, using training dataset for training and development dataset for training validation.
    For each example, we extracted a feature map using Oxford VGG16 \cite{simonyan2015vgg} pretrained on ImageNet \cite{russakovsky2015imagenet} with fixed weights.
    Then we applied ground truth bounding boxes from the MSCOCO dataset to pool the object features out of the feature map.
    
    The IC system used a following function for input projection:
    \begin{equation}
        f(v_i) = W_{in}v_i
    \end{equation}
    The size of the projection, embeddings and LSTM were 512.
    We used batch size 100, learning rate 2e-3 and applied early stopping when the CIDEr \cite{vedantam2014cider} score on the development dataset did not improve for 10 epochs.
    
    We then looked at the learned embeddings and the input projection.
    To analyze the embeddings, we extracted a subset of 65 words that overlap with the MSCOCO object detection labels.
    We analyzed object vectors extracted from the instances in the development dataset.
    Since each object class contains a various number of object vectors, we decided to compute object centroids for each class as the class representation.
    We used t-SNE \cite{maaten2008tsne} for vector visualization.
    
\begin{figure*}[ht]
    \centering
    \begin{minipage}[b]{\linewidth}
        \centering
        \includegraphics[width=0.4\linewidth]{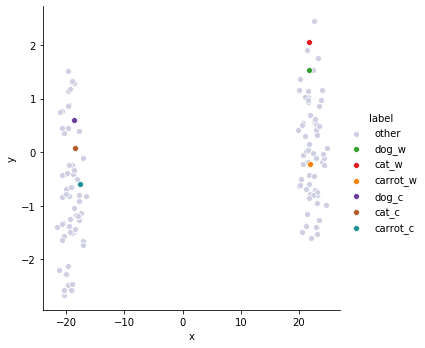}
        \includegraphics[width=0.4\linewidth]{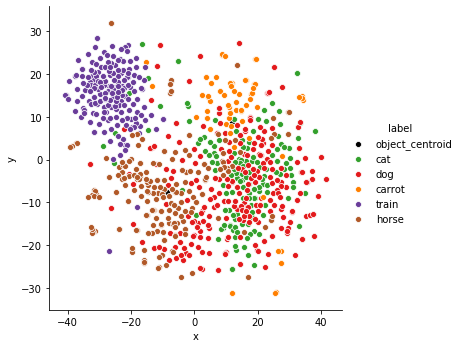}
    \end{minipage} 
    \begin{minipage}{\linewidth}
        \centering
        \includegraphics[width=0.4\linewidth]{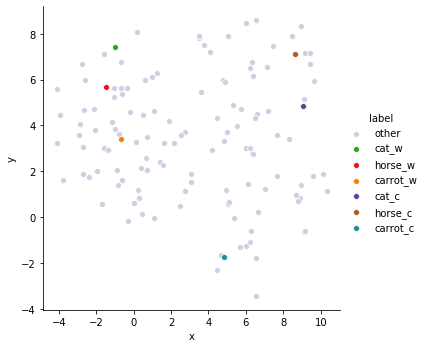}
        \includegraphics[width=0.4\linewidth]{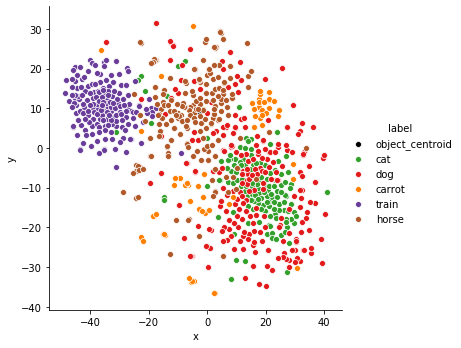}
    \end{minipage}
    \caption{A visualization of the distance between the object centroids and word embeddings (two figures in the left column) and overlap of the object vector clusters from various class (right column). The comparison is between the object representations/centroids before linear projection (top half) and after applying learned linear projection (bottom half). The projection and the word embeddings were extracted by our initial IC model.}
    \label{fig:initial-analysis}
\end{figure*}
    
    Figure~\ref{fig:initial-analysis} shows a proximity comparison between the object centroid space to the word embedding space (left) and a cluster overlap of the class vectors for a chosen subset of classes (right).
    We compare the VORs before the linear projection (top) and after its application.
    
    We can see, that without any implicit constraints, the IC system tries to learn a mapping that closes the distance between the object representations and word embeddings.
    However, both representations differ in the way they model relationships between the classes (e.g. carrot being close to the animals in the centroid cluster and distant in the word embedding cluster when projected by t-SNE).
    Furthermore, we can see that the learned input projection is unable to explicitly separate clusters of objects with a same class label.
    We hypothesise that by explicitly learning to separate these clusters we can improve the model performance.
    
    Based on these observations, we decided to apply an idea presented by \citet{bordes2019groundedspace} and ground the projected VOR space in the word embedding space using their grounding objectives.
    Similarly, we constrain our input projection function $f(v_i, \theta{}_f)$ by an additional grounding objective:
    \begin{equation}
        L_g(\theta{}_g) = \alpha{}_C L_C(\theta{}_f) + \alpha{}_P L_P(\theta{}_f)
    \end{equation}
    
    where $L_C$ and $L_P$ are the losses introduced in the original paper, $\alpha{}_C$ and $\alpha{}_P$ are the loss-weighting parameters and $\theta{}_f \subset \theta{}$, $\theta$ being trainable parameters of the whole network.
    
    The cluster information loss $L_C$ is a max-margin ranking loss \cite{karpathy2015deep, carvalho2018crossmodal} with the following formulation:
    \begin{equation}
        L_C = \frac{1}{N}\sum_{\mathclap{\substack{(z_i, z_j, z_k)\\
                              c(z_i) = c(z_j)\\
                              c(z_i) \neq c(z_k)}}} \lfloor{\gamma - cos(z_i, z_j) + cos(z_i, z_k)}\rfloor_+
    \end{equation}
    
    where $c(\cdot)$ denotes a class of the object vector.
    Due to the memory constraints, in practice we sample a subset of triplets $(z_i, z_j, z_k)$ of size $N$ from a training batch and normalize the loss sum by the number of valid triplets per sample.
    Using this loss we aim to explicitly separate clusters of object features according to their classes and examine what effect it has on the learning abilities of the image captioning system.
    
    The cluster loss alone does not provide any direct basis for the grounding in the textual modality.
    Our hypothesis is that a separation of VORs with different class labels without any structure can hurt the performance since it disregards actual relations between the classes.
    For example in Figure~\ref{fig:initial-analysis}, even though we want the representations of \textit{dog}, \textit{cat} and \textit{train} to be in separate cluster, we still want most of the time to have \textit{dog} closer to \textit{cat} rather then the \textit{train} because they are semantically closer.
    
    The perceptual information loss $L_P$ aims to solve this problem by modeling relations between the object representations based on the relations of the word embeddings of their labels. \citet{bordes2019groundedspace} express these relations through vector similarities, formulating the following loss:
    \begin{equation}
        L_P = -\rho(\{sim^{obj}_{i, j}\}, \{sim^{text}_{i, j}\})
    \end{equation}
    where $\rho$ is Pearson correlation. The similarity functions $sim^{obj}_{i, j}$ and $sim^{text}_{i, j}$ are defined as follows:
    \begin{equation}
        \begin{array}{l}
        sim^{obj}_{i, j} = cos(z_i, z_j) \\
        sim^{text}_{i, j} = cos(W_e\phi(c(z_i)), W_e\phi(c(z_j)))
        \end{array}
    \end{equation}
    
    Similar to cluster loss, we sample a subset of object vectors from the batch.
    Note that we compute similarity correlation only using vector pairs that belong to different classes to avoid further clustering this process might lead to.
    
    
    

\section{Experiments}
\label{sec:experiments}

\subsection{Experiment setup}

Similar to our initial run, we use MSCOCO dataset in our experiments with the Karpathy splits.
During training, we use CIDEr score on the validation split (5,000 images) as the early stopping criterion and we report results measured on the test split (another 5,000 images).
We use a default caption preprocessing available at the NeuralTalk2 \cite{karpathy2015deep} codebase\footnote{\url{https://github.com/karpathy/neuraltalk2}}, resulting in a captioning vocabulary of size around 9570 tokens.

In all our following experiments, we use the following network parameters:
\begin{itemize}
    \item batch size: 100
    \item optimizer: Adam \cite{kingma2015adam}
    \item initial learning rate: 2e-3
    \item exponential decay: 0.8, every 6000 steps
    \item gradient clipping (by norm): 1.0
    \item dropout: 0.2
    \item maximum sentence length: 16
\end{itemize}
When we compute the $L_C$ and $L_P$, we sample $N=500$ object vector instances from a current batch.
During training, we use teacher forcing \cite{sutskever2014seq2seq} where at each time step we use a ground truth token as an input instead of a previous network output.
We use greedy decoding during inference, outputting a token with the highest probability at each time step.
This is to remove any other variables from our experiments.
In practice, sampling techniques like beam search are commonly used to avoid phenomena such as exposure bias \cite{bengio2015scheduled, ranzato2016sequencelevel}.

In our experiments, we investigate two scenarios based on the process of the object feature extraction, \textit{CNN+SSD} and \textit{Bottom\_Up+SSD}.

\subparagraph{CNN+SSD.}
We use a Single Shot Multibox Detector (SSD) network available at Tensorflow Object Detection API \cite{huang2017tfodapi}.\footnote{\url{https://github.com/tensorflow/models/tree/master/research/object_detection}}
We decided to use the \textit{ssd\_mobilenet\_v2\_coco} model pretrained on the MSCOCO dataset mainly because of its fast performance.
The object detector output consists of object bounding boxes and their labels, however, it does not provide us with any explicit object representation vectors.
Therefore, we use the bounding boxes to extract object representations from the output of VGG16 \cite{simonyan2015vgg} similar to our initial experiment. We use the 14x14x512 feature map extracted from the last convolutional layer before applying max pooling. Then, we apply RoI pooling on the feature map using the bounding boxes to extract the VORs.

\begin{table*}[ht]
\begin{center}
\begin{tabular}{l|ccccc|c}
\toprule
 & BLEU-1 & BLEU-2 & BLEU-3 & BLEU-4 & ROUGE-L & CIDEr \\
\midrule
$CNN+SSD$ & 64.6 & 46.0 & 31.8 & 22.0 & 47.1 & 68.9 \\
\midrule
$+ L_C$ & \textbf{65.0} & \textbf{46.1} & \textbf{32.0} & \textbf{22.3} & 47.0 & \textbf{69.4} \\
$+ L_P$ & 63.2 & 44.1 & 29.8 & 20.1 & 45.6 & 65.5 \\
$+ L_C + L_P$ & 64.5 & 45.4 & 31.1 & 21.4 & 46.6 & 68.5 \\
\midrule
$Bottom\_Up+SSD$ & 65.0 & 46.4 & 32.3 & 22.4 & 47.1 & 70.5 \\
\midrule
$+ L_C$ & 64.6 & 46.0 & 31.9 & 22.1 & 46.9 & 68.7 \\
$+ L_P$ & 64.5 & 45.9 & 31.8 & 22.0 & 46.6 & 68.5 \\
$+ L_C + L_P$ & 64.6 & 46.2 & 32.2 & 22.3 & \textbf{47.2} & 68.8 \\
\midrule
$One\mhyphen{}Hot (SSD)$ & 63.3 & 43.9 & 29.5 & 20.3 & 45.6 & 66.2 \\
\bottomrule
\end{tabular}
\caption{Comparison of the performance between the suggested model baselines (\textit{CNN-SSD}, \textit{Bottom\_Up+SSD}) and their combination with the cluster loss ($L_C$), perceptual loss ($L_P$) or their combination ($L_C + L_P$). We include \textit{One-Hot} model for comparison. Results outperforming their respective baseline are in \textbf{bold}.}
\label{tab:final-results}
\end{center}
\end{table*}

\subparagraph{Bottom\_Up+SSD.}
We use pretrained object features provided by \citet{anderson2017up-down}.\footnote{\url{https://github.com/peteanderson80/Up-Down-Captioner}}
They extract the features using Faster RCNN network initialized with ResNet-101 \cite{he2016resnet} pretrained on ImageNet \cite{russakovsky2015imagenet} which they fine tune on the Visual Genome \cite{krishna2016visualgenome} dataset.
They further modify the object classifier to also predict attribute classes for every detected object.
The Visual Genome dataset contains a very large number of object and attribute classes, however, the authors use only a subset of 1600 object classes and 400 attribute classes for the model training.

The downside of using these pretrained features is a fact that the authors provide only the VORs and their bounding box information, however, they do not extract any object nor attribute labels.
To assign the MSCOCO labels to the extracted VORs we compute an intersection-over-union (IoU) between the pretrained representation bounding boxes and the bounding boxes of the objects extracted by the SSD network we used in the \textit{CNN+SSD} scenario.
Each object representation then gets assigned a label with the highest IoU value. If there is an IoU overlap of 0 with every SSD bounding box, we assign an \textit{UNK} label to the VOR vector.

In both scenarios, we provide a comparison with a baseline setting which is only optimized for the standard captioning cross-entropy objective.
It optimizes the trainable parameters $\theta$ using the ground truth caption sequence $y^{*}_{1:T}$:
\begin{equation}
    L_{XE} = \frac{1}{T}\sum_{t=1}^{T}log(p_\theta(y^{*}|y^{*}_{1:t-1}))
\end{equation}
We normalize the sum by the caption length to simplify its combination with the grounding losses.
We then compare a model variations trained using $L_{XE} + L_C$, $L_{XE} + L_P$ and $L_{XE} + L_C + L_P$ with the baseline. For the loss parameters, we set $\gamma{} = 0.5$ and $\alpha{}_C = \alpha{}_P = 1.0$.

For contrast, we also compare our results with a \textit{One-Hot} model which only uses one-hot label representations extracted by the SSD object detector.

\begin{table*}[ht]
\begin{center}
\begin{tabular}{l|c|cc|cc}
\toprule
 & CIDEr & mNNO & $\rho{}_{vis}$ & $C_{inter}$ & $C_{intra}$ \\ 
\midrule
$CNN+SSD$ & 68.9 & 24.6 / 31.7 & 2.5 / 6.0 & 1.0 (85.8) & 6.8 (87.5) \\
\midrule
$+ L_C$ & 69.4 & 31.2 / 32.3 & 4.5 / -2.3 & 23.8 & 43.4 \\
$+ L_P$ & 65.5 & \textbf{32.8} / \textbf{41.0} & 15.2 / 26.2 & 75.2 & \textbf{79.6} \\
$+ L_C + L_P$ & 68.5 & 31.2 / 36.9 & \textbf{15.5} / \textbf{53.1} & 42.0 & 62.3 \\
\midrule
$Bottom\_Up+SSD$ & 70.5 & 28.7 / 35.8 & 0.7 / 5.3 & 0.7 (86.5) & 3.7 (87.1) \\
\midrule
$+ L_C$ & 68.7 & 26.1 / 31.2 & 0.7 / 8.0 & 25.3 & 48.9 \\
$+ L_P$ & 68.5 & \textbf{32.8} / \textbf{37.9} & 6.1 / 10.6 & 86.5 & \textbf{87.6}  \\
$+ L_C + L_P$ & 68.8 & 31.2 / 34.8 & \textbf{6.4} / \textbf{32.1} & 67.0 & 77.2 \\
\bottomrule
\end{tabular}
\caption{Analysis of the structure of the learned projection space. mNNO and $\rho_{vis}$ compare the  \textit{original/projected} object embedding space. Values in brackets ($C_{inter}$, $C_{intra}$) are related to the original non-projected object spaces. The best results outperforming the baseline are in \textbf{bold}. We include their respective CIDEr scores for comparison.}
\label{tab:final-analysis}
\end{center}
\end{table*}

\subsection{Results}

Beside CIDEr, we used BLEU \cite{papineni2002bleu} and ROUGE-L \cite{lin2004rouge} metrics that are commonly used in the IC literature. We used the same implementation as in NeuralTalk2.\footnote{\url{https://github.com/tylin/coco-caption}}

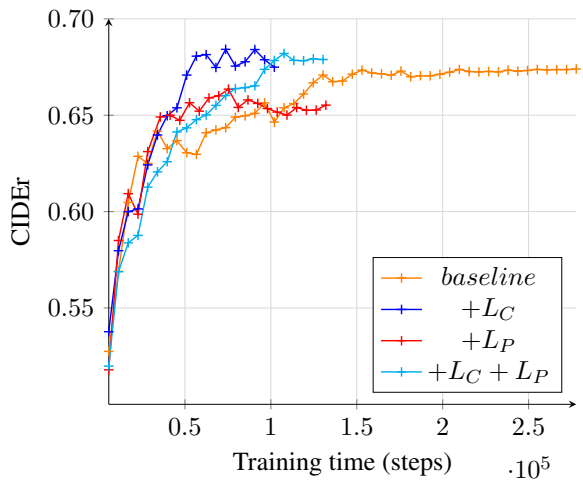
\begin{figure}
    \centering
    \def\verticalstep{0.8}

\begin{tikzpicture}[]
  \begin{axis}[cycle list name=tb, 
               grid=both,
               grid style={solid,gray!30!white},
               axis lines=middle,
               xlabel={Training time (steps)},
               ylabel={CIDEr},
               ymin=0.5,
               ymax=0.7,
               x label style={
                    at={(axis description cs:0.5,-0.1)},
                    anchor=north},
               y label style={
                    at={(axis description cs:-0.15,.5)},
                    rotate=90,
                    anchor=south},
               yticklabel style={%
                    /pgf/number format/.cd,
                    fixed,
                    fixed zerofill,
                    precision=2
                },
               legend pos=south east]
    \addplot table [x=Step, y=Value, col sep=comma] {./data/run-vgg16_bboxes-tag-epoch-CIDEr.csv};
    \addplot table [x=Step, y=Value, col sep=comma] {./data/run-vgg16_bboxes.final.cluster-tag-epoch-CIDEr.csv};
    \addplot table [x=Step, y=Value, col sep=comma] {./data/run-vgg16_bboxes.perceptual-tag-epoch-CIDEr.csv};
    \addplot table [x=Step, y=Value, col sep=comma] {./data/run-vgg16_bboxes.final.combined-tag-epoch-CIDEr.csv};
    \addlegendentry{$baseline$}
    \addlegendentry{$+L_C$}
    \addlegendentry{$+L_P$}
    \addlegendentry{$+L_C+L_P$}
  \end{axis}
\end{tikzpicture}
    \caption{Model performance on validation data during \textit{CNN+SSD} training showing faster convergence of the grounded models.}
    \label{fig:speed}
\end{figure}

Table~\ref{tab:final-results} shows a comparison between the described models.
For \textit{CNN+SSD}, we gained a small improvement using the clustering objective $L_C$.
That means that even objects cropped from a general feature map by a separately trained object detector can benefit from properly constrained input projection.
However, the perceptual loss seems to have a negative effect on the overall system performance, both alone ($L_P$) and leads to small drop in performance when combined with the cluster loss ($L_C + L_P$). However, Figure~\ref{fig:speed} shows that the grounded models reach the early stopping criterion about two to three times faster making their training much efficient.

The results are worse for the \textit{Bottom\_Up+SSD} where an additional training losses did not have any positive effect at all.
We suspect that the main reason might be mislabeling of the pretrained object vectors because we used an SSD with lower rate of detection together with a portion of VOR vectors being assigned the generic \textit{UNK} label.


Following \citet{bordes2019groundedspace}, we also provide results of a structural analysis in Table~\ref{tab:final-analysis}.
Note that we compute the mean nearest neighbor overlap (mNNO) \cite{collell2018crossmodal} as a proportion neighbor overlap between the word embeddings and object feature centroids of clusters for each MSCOCO label.
Other metrics are the same as the original paper.

Surprisingly, it seems that cluster homogeneity or a distinction between clusters is not crucial for the IC task.
The unconstrained model seems to learn structure regardless of the vector clustering.
On the other hand, the structure of the projected object vectors (mNNO) are always closer to the structure of the word embeddings regardless of whether we use the grounding losses or not.

The cluster loss was able to efficiently separate the clusters with different labels ($C_{inter}=23.8$ and $25.3$ in both scenarios) while still maintaining some cluster homogeneity ($C_{intra}=43.4$ and $48.9$ vs the $C_{intra}=6.8$ and $3.7$ in the baselines).

The perceptual loss managed to achieve the best results regarding the mNNO between the centroids and word embeddings while disabling the default unclustering behaviour of the baseline. This inability to separate the initial clusters could have been the reason for the lower CIDEr scores.
The combination of both losses then led to the best structural similarity correlation ($\rho{}_{vis}$).
Interesting observation is the overall increase in mNNO and visual correlation between the original VOR vectors and the learned word embeddings when the grounding losses are applied.
This suggests that the grounding between the modalities is mutual: not only the VOR vectors are being grounded in the textual modality, the word embeddings are also being grounded in the projected VOR space.

The results of the structural analysis are similar between \textit{CNN-SSD} and \textit{Bottom\_Up-SSD} while being slightly worse for the latter in the mNNO and $\rho{}_{vis}$.
Again, this is probably due to a sub-optimal label assignment for the pretrained bottom-up features.
    
\section{Conclusion}
\label{sec:conclusion}    

We presented a method for grounding of object feature vectors extracted by an object detection network in a word embedding space learned by the IC system and investigated a hypothesis whether such grounding can improve said IC system performance.
We used previously proposed grounding losses to enforce clustering of the object features based on their class label and a distance correlation loss to enforce a structural similarity between these features and corresponding label embeddings.

The results indicate that while having similar performance measured by the automatic metrics, the grounded models converge about two to three times faster than the original baseline.
Furthermore, the structural analysis showed that the unconstrained IC system completely ignores cluster information and is inclined to learn a projection that unclusters the object features regardless of their class.
Finally, the proposed grounding methods enforce a mutual grounding between the projected object features and the word embedding space.


In future, we plan to confirm the results with a more descriptive set of object features properly trained on a larger set of object labels such as Visual Genome, to rule-out a possibility that our results were biased by the coarseness of the relatively small set of labels in MSCOCO dataset. We also plan to further investigate the mutual grounding using different layers of textual representation, e.g. intermediate or output states of the caption decoder. 



\bibliography{acl2020.bib}

\begin{thebibliography}{45}
\expandafter\ifx\csname natexlab\endcsname\relax\def\natexlab#1{#1}\fi

\bibitem[{Anderson et~al.(2018)Anderson, He, Buehler, Teney, Johnson, Gould,
  and Zhang}]{anderson2017up-down}
Peter Anderson, Xiaodong He, Chris Buehler, Damien Teney, Mark Johnson, Stephen
  Gould, and Lei Zhang. 2018.
\newblock {Bottom-Up and Top-Down Attention for Image Captioning and Visual
  Question Answering}.
\newblock In \emph{CVPR}.

\bibitem[{Artetxe et~al.(2016)Artetxe, Labaka, and
  Agirre}]{artetxe2016learning}
Mikel Artetxe, Gorka Labaka, and Eneko Agirre. 2016.
\newblock \href {https://doi.org/10.18653/v1/D16-1250} {Learning principled
  bilingual mappings of word embeddings while preserving monolingual
  invariance}.
\newblock In \emph{Proceedings of the 2016 Conference on Empirical Methods in
  Natural Language Processing}, pages 2289--2294, Austin, Texas. Association
  for Computational Linguistics.

\bibitem[{Artetxe et~al.(2018)Artetxe, Labaka, and
  Agirre}]{artetxe2018unsupervised}
Mikel Artetxe, Gorka Labaka, and Eneko Agirre. 2018.
\newblock \href {https://doi.org/10.18653/v1/D18-1399} {Unsupervised
  statistical machine translation}.
\newblock In \emph{Proceedings of the 2018 Conference on Empirical Methods in
  Natural Language Processing}, pages 3632--3642, Brussels, Belgium.
  Association for Computational Linguistics.

\bibitem[{Bahdanau et~al.(2015)Bahdanau, Cho, and
  Bengio}]{bahdanau2015attention}
Dzmitry Bahdanau, Kyunghyun Cho, and Yoshua Bengio. 2015.
\newblock \href {http://arxiv.org/abs/1409.0473} {{Neural Machine Translation
  by Jointly Learning to Align and Translate}}.
\newblock In  \cite{DBLP:conf/iclr/2015}.

\bibitem[{Bengio et~al.(2015)Bengio, Vinyals, Jaitly, and
  Shazeer}]{bengio2015scheduled}
Samy Bengio, Oriol Vinyals, Navdeep Jaitly, and Noam Shazeer. 2015.
\newblock \href {http://dl.acm.org/citation.cfm?id=2969239.2969370} {Scheduled
  sampling for sequence prediction with recurrent neural networks}.
\newblock In \emph{Proceedings of the 28th International Conference on Neural
  Information Processing Systems - Volume 1}, NIPS'15, pages 1171--1179,
  Cambridge, MA, USA. MIT Press.

\bibitem[{Bengio and LeCun(2015)}]{DBLP:conf/iclr/2015}
Yoshua Bengio and Yann LeCun, editors. 2015.
\newblock \href
  {https://iclr.cc/archive/www/doku.php\%3Fid=iclr2015:accepted-main.html}
  {\emph{{3rd International Conference on Learning Representations, {ICLR}
  2015, San Diego, CA, USA, May 7-9, 2015, Conference Track Proceedings}}}.

\bibitem[{Bordes et~al.(2019)Bordes, Zablocki, Soulier, Piwowarski, and
  Gallinari}]{bordes2019groundedspace}
Patrick Bordes, Eloi Zablocki, Laure Soulier, Benjamin Piwowarski, and patrick
  Gallinari. 2019.
\newblock \href {https://doi.org/10.18653/v1/D19-1064} {{Incorporating Visual
  Semantics into Sentence Representations within a Grounded Space}}.
\newblock In \emph{Proceedings of the 2019 Conference on Empirical Methods in
  Natural Language Processing and the 9th International Joint Conference on
  Natural Language Processing (EMNLP-IJCNLP)}, pages 696--707, Hong Kong,
  China. Association for Computational Linguistics.

\bibitem[{Carvalho et~al.(2018)Carvalho, Cad\`{e}ne, Picard, Soulier, Thome,
  and Cord}]{carvalho2018crossmodal}
Micael Carvalho, R{\'e}mi Cad\`{e}ne, David Picard, Laure Soulier, Nicolas
  Thome, and Matthieu Cord. 2018.
\newblock \href {https://doi.org/10.1145/3209978.3210036} {{Cross-Modal
  Retrieval in the Cooking Context: Learning Semantic Text-Image Embeddings}}.
\newblock In \emph{The 41st International ACM SIGIR Conference on Research \&
  Development in Information Retrieval}, SIGIR '18, pages 35--44, New York, NY,
  USA. ACM.

\bibitem[{Chrupa{\l}a et~al.(2015)Chrupa{\l}a, K{\'a}d{\'a}r, and
  Alishahi}]{chrupala2015learning}
Grzegorz Chrupa{\l}a, {\'A}kos K{\'a}d{\'a}r, and Afra Alishahi. 2015.
\newblock \href {https://doi.org/10.3115/v1/P15-2019} {{Learning Language
  through Pictures}}.
\newblock In \emph{Proceedings of the 53rd Annual Meeting of the Association
  for Computational Linguistics and the 7th International Joint Conference on
  Natural Language Processing (Volume 2: Short Papers)}, pages 112--118,
  Beijing, China. Association for Computational Linguistics.

\bibitem[{Collell and Moens(2018)}]{collell2018crossmodal}
Guillem Collell and Marie-Francine Moens. 2018.
\newblock \href {https://doi.org/10.18653/v1/P18-2074} {{Do Neural Network
  Cross-Modal Mappings Really Bridge Modalities?}}
\newblock In \emph{Proceedings of the 56th Annual Meeting of the Association
  for Computational Linguistics (Volume 2: Short Papers)}, pages 462--468,
  Melbourne, Australia. Association for Computational Linguistics.

\bibitem[{Collell et~al.(2017)Collell, Zhang, and Moens}]{collell2017imagined}
Guillem Collell, Ted Zhang, and Marie-Francine Moens. 2017.
\newblock \href {http://dl.acm.org/citation.cfm?id=3298023.3298203} {{Imagined
  Visual Representations As Multimodal Embeddings}}.
\newblock In \emph{Proceedings of the Thirty-First AAAI Conference on
  Artificial Intelligence}, AAAI'17, pages 4378--4384. AAAI Press.

\bibitem[{Girshick et~al.(2014)Girshick, Donahue, Darrell, and
  Malik}]{girshick2014rcnn}
Ross Girshick, Jeff Donahue, Trevor Darrell, and Jitendra Malik. 2014.
\newblock \href {https://doi.org/10.1109/CVPR.2014.81} {{Rich Feature
  Hierarchies for Accurate Object Detection and Semantic Segmentation}}.
\newblock In \emph{Proceedings of the 2014 IEEE Conference on Computer Vision
  and Pattern Recognition}, CVPR '14, pages 580--587, Washington, DC, USA. IEEE
  Computer Society.

\bibitem[{Gordon and Van~Durme(2013)}]{gordon2013textbias}
Jonathan Gordon and Benjamin Van~Durme. 2013.
\newblock \href {https://doi.org/10.1145/2509558.2509563} {{Reporting Bias and
  Knowledge Acquisition}}.
\newblock In \emph{Proceedings of the 2013 Workshop on Automated Knowledge Base
  Construction}, AKBC '13, pages 25--30, New York, NY, USA. ACM.

\bibitem[{Guo et~al.(2019)Guo, Wang, and Wang}]{guo2019replearningsurvey}
Wenzhong Guo, Jianwen Wang, and Shiping Wang. 2019.
\newblock \href {https://doi.org/10.1109/ACCESS.2019.2916887} {{Deep Multimodal
  Representation Learning: {A} Survey}}.
\newblock \emph{{IEEE} Access}, 7:63373--63394.

\bibitem[{{He} et~al.(2016){He}, {Zhang}, {Ren}, and {Sun}}]{he2016resnet}
K.~{He}, X.~{Zhang}, S.~{Ren}, and J.~{Sun}. 2016.
\newblock \href {https://doi.org/10.1109/CVPR.2016.90} {{Deep Residual Learning
  for Image Recognition}}.
\newblock In \emph{2016 IEEE Conference on Computer Vision and Pattern
  Recognition (CVPR)}, pages 770--778.

\bibitem[{Hill et~al.(2016)Hill, Cho, and Korhonen}]{hill2016learning}
Felix Hill, Kyunghyun Cho, and Anna Korhonen. 2016.
\newblock \href {https://doi.org/10.18653/v1/N16-1162} {{Learning Distributed
  Representations of Sentences from Unlabelled Data}}.
\newblock In \emph{Proceedings of the 2016 Conference of the North {A}merican
  Chapter of the Association for Computational Linguistics: Human Language
  Technologies}, pages 1367--1377, San Diego, California. Association for
  Computational Linguistics.

\bibitem[{Hochreiter and Schmidhuber(1997)}]{hochreiter1997lstm}
Sepp Hochreiter and J\"{u}rgen Schmidhuber. 1997.
\newblock \href {https://doi.org/10.1162/neco.1997.9.8.1735} {{Long Short-Term
  Memory}}.
\newblock \emph{Neural Comput.}, 9(8):1735--1780.

\bibitem[{{Huang} et~al.(2017){Huang}, {Rathod}, {Sun}, {Zhu}, {Korattikara},
  {Fathi}, {Fischer}, {Wojna}, {Song}, {Guadarrama}, and
  {Murphy}}]{huang2017tfodapi}
J.~{Huang}, V.~{Rathod}, C.~{Sun}, M.~{Zhu}, A.~{Korattikara}, A.~{Fathi},
  I.~{Fischer}, Z.~{Wojna}, Y.~{Song}, S.~{Guadarrama}, and K.~{Murphy}. 2017.
\newblock \href {https://doi.org/10.1109/CVPR.2017.351} {{Speed/Accuracy
  Trade-Offs for Modern Convolutional Object Detectors}}.
\newblock In \emph{2017 IEEE Conference on Computer Vision and Pattern
  Recognition (CVPR)}, pages 3296--3297.

\bibitem[{Karpathy and Li(2015)}]{karpathy2015deep}
Andrej Karpathy and Fei{-}Fei Li. 2015.
\newblock \href {https://doi.org/10.1109/CVPR.2015.7298932} {{Deep
  visual-semantic alignments for generating image descriptions}}.
\newblock In \emph{{IEEE} Conference on Computer Vision and Pattern
  Recognition, {CVPR} 2015, Boston, MA, USA, June 7-12, 2015}, pages
  3128--3137. {IEEE} Computer Society.

\bibitem[{Kiela et~al.(2018)Kiela, Conneau, Jabri, and
  Nickel}]{kiela2018learning}
Douwe Kiela, Alexis Conneau, Allan Jabri, and Maximilian Nickel. 2018.
\newblock \href {https://doi.org/10.18653/v1/N18-1038} {{Learning Visually
  Grounded Sentence Representations}}.
\newblock In \emph{Proceedings of the 2018 Conference of the North {A}merican
  Chapter of the Association for Computational Linguistics: Human Language
  Technologies, Volume 1 (Long Papers)}, pages 408--418, New Orleans,
  Louisiana. Association for Computational Linguistics.

\bibitem[{Kingma and Ba(2015)}]{kingma2015adam}
Diederik~P. Kingma and Jimmy Ba. 2015.
\newblock \href {http://arxiv.org/abs/1412.6980} {{Adam: {A} Method for
  Stochastic Optimization}}.
\newblock In  \cite{DBLP:conf/iclr/2015}.

\bibitem[{Krishna et~al.(2017)Krishna, Zhu, Groth, Johnson, Hata, Kravitz,
  Chen, Kalantidis, Li, Shamma, Bernstein, and
  Fei-Fei}]{krishna2016visualgenome}
Ranjay Krishna, Yuke Zhu, Oliver Groth, Justin Johnson, Kenji Hata, Joshua
  Kravitz, Stephanie Chen, Yannis Kalantidis, Li-Jia Li, David~A. Shamma,
  Michael~S. Bernstein, and Li~Fei-Fei. 2017.
\newblock \href {https://doi.org/10.1007/s11263-016-0981-7} {{Visual Genome:
  Connecting Language and Vision Using Crowdsourced Dense Image Annotations}}.
\newblock \emph{Int. J. Comput. Vision}, 123(1):32--73.

\bibitem[{Lazaridou et~al.(2015)Lazaridou, Pham, and
  Baroni}]{lazaridou2015combining}
Angeliki Lazaridou, Nghia~The Pham, and Marco Baroni. 2015.
\newblock \href {https://doi.org/10.3115/v1/N15-1016} {{Combining Language and
  Vision with a Multimodal Skip-gram Model}}.
\newblock In \emph{Proceedings of the 2015 Conference of the North {A}merican
  Chapter of the Association for Computational Linguistics: Human Language
  Technologies}, pages 153--163, Denver, Colorado. Association for
  Computational Linguistics.

\bibitem[{Lin(2004)}]{lin2004rouge}
Chin-Yew Lin. 2004.
\newblock \href {https://www.aclweb.org/anthology/W04-1013} {{ROUGE}: A package
  for automatic evaluation of summaries}.
\newblock In \emph{Text Summarization Branches Out}, pages 74--81, Barcelona,
  Spain. Association for Computational Linguistics.

\bibitem[{Lin et~al.(2014)Lin, Maire, Belongie, Hays, Perona, Ramanan,
  Doll{\'a}r, and Zitnick}]{Lin2014mscoco}
Tsung-Yi Lin, Michael Maire, Serge Belongie, James Hays, Pietro Perona, Deva
  Ramanan, Piotr Doll{\'a}r, and C.~Lawrence Zitnick. 2014.
\newblock {Microsoft COCO: Common Objects in Context}.
\newblock In \emph{Computer Vision -- ECCV 2014}, pages 740--755, Cham.
  Springer International Publishing.

\bibitem[{Luong et~al.(2015)Luong, Pham, and Manning}]{luong2015attention}
Thang Luong, Hieu Pham, and Christopher~D. Manning. 2015.
\newblock \href {https://doi.org/10.18653/v1/D15-1166} {{Effective Approaches
  to Attention-based Neural Machine Translation}}.
\newblock In \emph{Proceedings of the 2015 Conference on Empirical Methods in
  Natural Language Processing}, pages 1412--1421, Lisbon, Portugal. Association
  for Computational Linguistics.

\bibitem[{Madhyastha et~al.(2018)Madhyastha, Wang, and
  Specia}]{madhyastha2018exploits}
Pranava~Swaroop Madhyastha, Josiah Wang, and Lucia Specia. 2018.
\newblock \href {http://bmvc2018.org/contents/papers/0925.pdf} {{End-to-end
  Image Captioning Exploits Distributional Similarity in Multimodal Space}}.
\newblock In \emph{British Machine Vision Conference 2018, {BMVC} 2018,
  Northumbria University, Newcastle, UK, September 3-6, 2018}, page 306. {BMVA}
  Press.

\bibitem[{Mikolov et~al.(2013{\natexlab{a}})Mikolov, Chen, Corrado, and
  Dean}]{mikolov2013efficient}
Tomas Mikolov, Kai Chen, Greg Corrado, and Jeffrey Dean. 2013{\natexlab{a}}.
\newblock \href {http://arxiv.org/abs/1301.3781} {{Efficient Estimation of Word
  Representations in Vector Space}}.
\newblock In \emph{1st International Conference on Learning Representations,
  {ICLR} 2013, Scottsdale, Arizona, USA, May 2-4, 2013, Workshop Track
  Proceedings}.

\bibitem[{Mikolov et~al.(2013{\natexlab{b}})Mikolov, Sutskever, Chen, Corrado,
  and Dean}]{mikolov2013word2vec}
Tomas Mikolov, Ilya Sutskever, Kai Chen, Greg Corrado, and Jeffrey Dean.
  2013{\natexlab{b}}.
\newblock \href {http://dl.acm.org/citation.cfm?id=2999792.2999959}
  {{Distributed Representations of Words and Phrases and Their
  Compositionality}}.
\newblock In \emph{Proceedings of the 26th International Conference on Neural
  Information Processing Systems - Volume 2}, NIPS'13, pages 3111--3119, USA.
  Curran Associates Inc.

\bibitem[{Papineni et~al.(2002)Papineni, Roukos, Ward, and
  Zhu}]{papineni2002bleu}
Kishore Papineni, Salim Roukos, Todd Ward, and Wei-Jing Zhu. 2002.
\newblock \href {https://doi.org/10.3115/1073083.1073135} {{B}leu: a method for
  automatic evaluation of machine translation}.
\newblock In \emph{Proceedings of the 40th Annual Meeting of the Association
  for Computational Linguistics}, pages 311--318, Philadelphia, Pennsylvania,
  USA. Association for Computational Linguistics.

\bibitem[{Plummer et~al.(2017)Plummer, Wang, Cervantes, Caicedo, Hockenmaier,
  and Lazebnik}]{plummer2017flickr30k}
Bryan~A. Plummer, Liwei Wang, Chris~M. Cervantes, Juan~C. Caicedo, Julia
  Hockenmaier, and Svetlana Lazebnik. 2017.
\newblock \href {https://doi.org/10.1007/s11263-016-0965-7} {{Flickr30K
  Entities: Collecting Region-to-Phrase Correspondences for Richer
  Image-to-Sentence Models}}.
\newblock \emph{Int. J. Comput. Vision}, 123(1):74--93.

\bibitem[{Ranzato et~al.(2016)Ranzato, Chopra, Auli, and
  Zaremba}]{ranzato2016sequencelevel}
Marc'Aurelio Ranzato, Sumit Chopra, Michael Auli, and Wojciech Zaremba. 2016.
\newblock \href {http://arxiv.org/abs/1511.06732} {Sequence level training with
  recurrent neural networks}.
\newblock In \emph{4th International Conference on Learning Representations,
  {ICLR} 2016, San Juan, Puerto Rico, May 2-4, 2016, Conference Track
  Proceedings}.

\bibitem[{Redmon et~al.(2016)Redmon, Divvala, Girshick, and
  Farhadi}]{redmon2016yolo}
Joseph Redmon, Santosh~Kumar Divvala, Ross~B. Girshick, and Ali Farhadi. 2016.
\newblock \href {https://doi.org/10.1109/CVPR.2016.91} {{You Only Look Once:
  Unified, Real-Time Object Detection}}.
\newblock In \emph{2016 {IEEE} Conference on Computer Vision and Pattern
  Recognition, {CVPR} 2016, Las Vegas, NV, USA, June 27-30, 2016}, pages
  779--788. {IEEE} Computer Society.

\bibitem[{Ren et~al.(2015)Ren, He, Girshick, and Sun}]{shaoqing2015fasterrcnn}
Shaoqing Ren, Kaiming He, Ross Girshick, and Jian Sun. 2015.
\newblock \href {http://dl.acm.org/citation.cfm?id=2969239.2969250} {{Faster
  R-CNN: Towards Real-time Object Detection with Region Proposal Networks}}.
\newblock In \emph{Proceedings of the 28th International Conference on Neural
  Information Processing Systems - Volume 1}, NIPS'15, pages 91--99, Cambridge,
  MA, USA. MIT Press.

\bibitem[{Russakovsky et~al.(2015)Russakovsky, Deng, Su, Krause, Satheesh, Ma,
  Huang, Karpathy, Khosla, Bernstein, Berg, and
  Fei-Fei}]{russakovsky2015imagenet}
Olga Russakovsky, Jia Deng, Hao Su, Jonathan Krause, Sanjeev Satheesh, Sean Ma,
  Zhiheng Huang, Andrej Karpathy, Aditya Khosla, Michael Bernstein,
  Alexander~C. Berg, and Li~Fei-Fei. 2015.
\newblock \href {https://doi.org/10.1007/s11263-015-0816-y} {{ImageNet Large
  Scale Visual Recognition Challenge}}.
\newblock \emph{Int. J. Comput. Vision}, 115(3):211--252.

\bibitem[{Silberer and Lapata(2014)}]{silberer2014learning}
Carina Silberer and Mirella Lapata. 2014.
\newblock \href {https://doi.org/10.3115/v1/P14-1068} {{Learning Grounded
  Meaning Representations with Autoencoders}}.
\newblock In \emph{Proceedings of the 52nd Annual Meeting of the Association
  for Computational Linguistics (Volume 1: Long Papers)}, pages 721--732,
  Baltimore, Maryland. Association for Computational Linguistics.

\bibitem[{Simonyan and Zisserman(2015)}]{simonyan2015vgg}
Karen Simonyan and Andrew Zisserman. 2015.
\newblock {Very Deep Convolutional Networks for Large-Scale Image Recognition}.
\newblock In \emph{International Conference on Learning Representations}.

\bibitem[{Sutskever et~al.(2014)Sutskever, Vinyals, and
  Le}]{sutskever2014seq2seq}
Ilya Sutskever, Oriol Vinyals, and Quoc~V. Le. 2014.
\newblock \href {http://dl.acm.org/citation.cfm?id=2969033.2969173} {Sequence
  to sequence learning with neural networks}.
\newblock In \emph{Proceedings of the 27th International Conference on Neural
  Information Processing Systems - Volume 2}, NIPS'14, pages 3104--3112,
  Cambridge, MA, USA. MIT Press.

\bibitem[{{van der Maaten} and Hinton(2008)}]{maaten2008tsne}
L.J.P. {van der Maaten} and G.E. Hinton. 2008.
\newblock {Visualizing High-Dimensional Data Using t-SNE}.
\newblock \emph{Journal of Machine Learning Research}, 9:2579--2605.

\bibitem[{Vedantam et~al.(2014)Vedantam, Zitnick, and
  Parikh}]{vedantam2014cider}
Ramakrishna Vedantam, C.~Lawrence Zitnick, and Devi Parikh. 2014.
\newblock {CIDEr: Consensus-based image description evaluation}.
\newblock \emph{2015 IEEE Conference on Computer Vision and Pattern Recognition
  (CVPR)}, pages 4566--4575.

\bibitem[{Vinyals et~al.(2017)Vinyals, Toshev, Bengio, and
  Erhan}]{vinyals2015show}
Oriol Vinyals, Alexander Toshev, Samy Bengio, and Dumitru Erhan. 2017.
\newblock \href {https://doi.org/10.1109/TPAMI.2016.2587640} {{Show and Tell:
  Lessons Learned from the 2015 MSCOCO Image Captioning Challenge}}.
\newblock \emph{IEEE Trans. Pattern Anal. Mach. Intell.}, 39(4):652--663.

\bibitem[{Wang et~al.(2018)Wang, Madhyastha, and Specia}]{wang2018counts}
Josiah Wang, Pranava~Swaroop Madhyastha, and Lucia Specia. 2018.
\newblock \href {https://doi.org/10.18653/v1/N18-1198} {{Object Counts!
  Bringing Explicit Detections Back into Image Captioning}}.
\newblock In \emph{Proceedings of the 2018 Conference of the North {A}merican
  Chapter of the Association for Computational Linguistics: Human Language
  Technologies, Volume 1 (Long Papers)}, pages 2180--2193, New Orleans,
  Louisiana. Association for Computational Linguistics.

\bibitem[{Xu et~al.(2015)Xu, Ba, Kiros, Cho, Courville, Salakhudinov, Zemel,
  and Bengio}]{xu2015showattend}
Kelvin Xu, Jimmy Ba, Ryan Kiros, Kyunghyun Cho, Aaron Courville, Ruslan
  Salakhudinov, Rich Zemel, and Yoshua Bengio. 2015.
\newblock \href {http://proceedings.mlr.press/v37/xuc15.html} {{Show, Attend
  and Tell: Neural Image Caption Generation with Visual Attention}}.
\newblock In \emph{Proceedings of the 32nd International Conference on Machine
  Learning}, volume~37 of \emph{Proceedings of Machine Learning Research},
  pages 2048--2057, Lille, France. PMLR.

\bibitem[{Yin and Ordonez(2017)}]{yin2017obj2text}
Xuwang Yin and Vicente Ordonez. 2017.
\newblock \href {https://doi.org/10.18653/v1/D17-1017} {{{O}bj2{T}ext:
  Generating Visually Descriptive Language from Object Layouts}}.
\newblock In \emph{Proceedings of the 2017 Conference on Empirical Methods in
  Natural Language Processing}, pages 177--187, Copenhagen, Denmark.
  Association for Computational Linguistics.

\bibitem[{Zablocki et~al.(2018)Zablocki, Piwowarski, Soulier, and
  Gallinari}]{zablocki2018learning}
{\'E}loi Zablocki, Benjamin Piwowarski, Laure Soulier, and Patrick Gallinari.
  2018.
\newblock \href {https://hal.archives-ouvertes.fr/hal-01632414} {{Learning
  Multi-Modal Word Representation Grounded in Visual Context}}.
\newblock In \emph{{Association for the Advancement of Artificial Intelligence
  (AAAI)}}, New Orleans, United States.

\end{thebibliography}
\bibliographystyle{acl_natbib}



\end{document}